\documentclass{IOS-Book-Article}

\usepackage{mathptmx}
\usepackage{soul}\setuldepth{article}
\usepackage{microtype}      % microtypography
\usepackage{xcolor}         % colors

\usepackage{amsmath}
\usepackage{amssymb}
\usepackage{mathtools}
\usepackage{amsthm}
\usepackage{multirow}
\usepackage{array}
\usepackage{booktabs}
\usepackage{pifont}
\usepackage{tabularx}
\usepackage{array}
\newcolumntype{Y}{>{\raggedright\arraybackslash}X} % 定义左对齐弹性列

\def\hb{\hbox to 11.5 cm{}}

\begin{document}

\pagestyle{headings}
\def\thepage{}
\begin{frontmatter}              % The preamble begins here.

%\pretitle{Pretitle}
\title{An Electrocardiogram Multi-task Benchmark with Comprehensive Evaluations and Insightful Findings}

\markboth{}{December 2024\hb}
%\subtitle{Subtitle}

\author[A]{\fnms{Yuhao} \snm{Xu}}
\author[C]{\fnms{Jiaying} \snm{Lu}}
\author[B]{\fnms{Sirui} \snm{Ding}}
\author[D]{\fnms{Defu} \snm{Cao}}
\author[C]{\fnms{Xiao} \snm{Hu}}
and
\author[A]{{\fnms{Carl} \snm{Yang}}%\orcid{....-....-....-....}%
\thanks{Corresponding Author: Carl Yang, j.carlyang@emory.edu.\\ Accepted by The 20th World Congress on Medical and Health Informatics (MedInfo 2025)}}

\runningauthor{Yuhao Xu et al.}
\address[A]{Department of Computer Science, Emory University}
\address[B]{Bakar Computational Health Sciences Institute, University of California, San Francisco}
\address[C]{Center for Data Science, School of Nursing, Emory University}
\address[D]{Department of Computer Science, University of Southern California}

\begin{abstract}
In the process of patient diagnosis, non-invasive measurements are widely used due to their low risks and quick results. Electrocardiogram~(ECG), as a non-invasive method to collect heart activities, is used to diagnose cardiac conditions. Analyzing the ECG typically requires domain expertise, which is a roadblock to applying artificial intelligence~(AI) for healthcare. 
Through advances in self-supervised learning and foundation models, AI systems can now acquire and leverage domain knowledge without relying solely on human expertise.
%Driven by big data and self-supervised learning, domain knowledge can be acquired by AI through foundation models. 
However, there is a lack of comprehensive analyses over the foundation models' performance on ECG. This study aims to answer the research question: \textit{``Are Foundation Models Useful for
ECG Analysis?''} To address it, we evaluate language~/~general time-series~/~ECG foundation models in comparison with time-series deep learning models. The experimental results show that general time-series~/~ECG foundation models achieve a top performance rate of 80\%, indicating their effectiveness in ECG analysis. In-depth analyses and insights are provided along with comprehensive experimental results. This study highlights the limitations and potential of foundation models in advancing physiological waveform analysis. The data and code for this benchmark are publicly available at \url{https://github.com/yuhaoxu99/ECGMultitasks-Benchmark}.

% \href{https://github.com/yuhaoxu99/ECGMultitasks-Benchmark}
% {https://github.com/yuhaoxu99/ECGMultitasks-Benchmark}. 

\end{abstract}

\end{frontmatter}
\markboth{December 2024\hb}{December 2024\hb}

\section{Introduction}

Electrocardiogram~(ECG) records the heart's electrical activities via skin-placed electrodes~\cite{alghatrif2012brief}, producing waveforms that decipher cardiac functions. Its non-invasive nature and ease of collection make ECG ideal for continuous monitoring and early detection of cardiovascular abnormalities. ECG is used for diagnosing arrhythmias~\cite{goldberger2000physiobank}, myocardial infarctions~\cite{acharya2017application} and analyzing heart rate variability~\cite{parak2011ecg}, highlighting its diverse utilities. However, ECG analysis is challenging due to individual variations, complex waveforms, and susceptibility to noises~\cite{acharya2017deep}.
Traditional ECG analysis relies on specialized clinicians, which is resource-intensive and does not scale well with large data volumes, increasing the risk of diagnostic errors. Advances in artificial intelligence~(AI) have led to AI-assisted ECG diagnostics surpassing human performance~\cite{al2023machine}.

AI models enhance ECG analysis by extracting rich features. Beyond detecting heart diseases, ECG can infer age~\cite{attia2019age}, gender~\cite{attia2019age}, blood pressure~\cite{simjanoska2018non}, and potassium levels~\cite{von2024evaluating}. 
Al-Zaiti et al.~\cite{al2023machine} used a random forest model that outperformed clinicians and FDA-approved systems in detecting acute myocardial ischemia. 
While traditional machine learning relies on feature engineering, potentially losing clinically relevant information, neural networks can use raw ECG signals, preserving critical information.
Baloglu et al.~\cite{baloglu2019classification} achieved over 99\% accuracy in myocardial infarction detection using a convolutional neural network. However, neural networks require extensive labeled data, which may not always be available. Foundation models address this by leveraging large-scale pre-training and task-specific fine-tuning. McKeen et al.~\cite{mckeen2024ecg} proposed ECG-FM, a transformer-based model pre-trained on 2.5 million samples, demonstrating the strong potential of unsupervised foundation models.
%The limitations of traditional machine learning and the data demands of neural networks highlight the necessity of ECG foundation models. These models retain critical information through large-scale pre-training, reduce the need for extensive labeled data via transfer learning, enhance generalizability, and offer open-weight solutions, promoting reproducibility and innovation in ECG analysis. 
Despite the emergence of ECG foundation models, fair and comprehensive evaluations on their effectiveness are lacking.

In this study, we construct a benchmark to fairly evaluate existing foundation models for ECG analysis, including large language model~(LLM), time-series foundation model~(TSFM), and an ECG foundation model~(ECGFM), in contrast to traditional time-series deep learning model~(TSDL). We compare their performance across five tasks, assessing ECG data modeling from different perspectives: simple feature extraction (RR interval estimation), complex feature extraction (age estimation), balanced labels (gender classification), imbalanced labels (potassium abnormality prediction), and multi-class classification (arrhythmia detection). Our evaluation scenarios encompass zero-shot, few-shot, and fine-tuning approaches. Through these comparisons, we analyze the strengths and weaknesses of different models and explore the effectiveness of foundation models. We envision our findings can inspire advancements in using foundation models for physiological waveform analysis. Our code is open-sourced to support future research.

\section{Methods}

% Since ECG is a typical time-series signal, it can be analyzed using time-series foundation models. However, the existing ECG foundation models are usually designed and pre-trained specifically for ECG data rather than fine-tuning more general time-series foundation models to ECG data and downstream tasks. It remains unclear whether these more general time-series foundation models perform well on ECG data. Therefore, it is necessary to establish a fair and comprehensive benchmark to explore the effectiveness of foundation models on ECG data.

Our experiment is conducted on the MIMIC-IV-ECG~\cite{gow2023mimic} dataset, which is currently the largest publicly accessible ECG dataset, comprising 800,035 diagnostic electrocardiograms from 161,352 unique patients.
Each ECG strip is 12-lead and 10 seconds in length with 500 Hz sampling rate, denoted by $x \in \mathbb{R}^{C\times L}$ where $C=12$ and $L=10\times500=5000$.

% In this study, our benchmark is designed as follows: (1) five different tasks are included, consisting of regression tasks for RR interval and age prediction, as well as classification tasks for gender, blood potassium level, and arrhythmia detection; (2) it includes four types of models: time-series deep learning models, large language models, time-series foundation models, and ECG foundation models; (3) it incorporates three training methods: zero-shot, few-shot, and fine-tuning.

% \subsection{Data}

% The MIMIC-IV-ECG dataset is currently the largest publicly accessible ECG dataset, containing 800,035 diagnostic electrocardiograms from 161,352 unique patients. In the zero-shot setting, we select 10,000 patients, totaling approximately 50,000 samples, for the experiment. For the few-shot setting, we choose 64-shot, meaning fine-tuning is conducted on 64 patients, followed by testing on 10,000 patients. In the fine-tuning setting, we use data from 10,000 patients, split into training, validation, and testing with a ratio of 70\%:20\%:10\%. Each experiment is repeated three times with different patients selected each time.

% \subsection{Downstream Tasks}

\vspace{\baselineskip}

\noindent \textbf{Downstream tasks.} We evaluate the performance of the benchmark on the following tasks:
(1) \textbf{\textit{RR Interval Estimation.}} The RR interval, which represents the time between two R-wave peaks in an ECG, is directly calculated from the ECG signal.
(2) \textbf{\textit{Age Estimation.}} Patient age estimation involves analyzing ECG signal characteristics to estimate age, challenging the model to effectively interpret complex signal patterns correlated with physiological aging.
(3) \textbf{\textit{Gender Classification.}} Gender classification is a binary classification task with a roughly balanced ratio of 50\% to 50\%.
(4) \textbf{\textit{Potassium Abnormality Prediction.}} We use ECG strips to predict the Potassium (blood) lab test result which is taken between ECG recording time and one hour after the ECG time. This task is challengening, with imbalanced ratio of 97\% (normal) to 3\% (abnormal).
(5) \textbf{\textit{Arrhythmia Detection.}} We select the 14 most frequently occurring diagnoses, with the remaining ones grouped under ``Others'', resulting in a total of 15 labels.
Among these downstream tasks, RR interval estimation and age estimation are regression tasks, where the prediction target $y \in \mathbb{R}$.
Gender prediction and potassium abnormality prediction are binary classification tasks, where the prediction target $y \in \{0, 1\}$. Arrhythmia detection is multiclass classification task, where the prediction target $y \in \{1, 2, 3, \dots, M\}$ ($M=15$ denotes the phenotype of arrhythmia). 
% For each downstream task, the goal of to train a model is to predict the targets using the input ECG strips: $Y_{tst}=f_\theta(X_{tst})$, where $f_\theta$ denotes a model $f$ with learnable parameter $\theta$ trained in the training set $\{X_{trn}, Y_{trn}\}$.

%These two tasks help evaluate whether the model accurately captures the relevant features in regression tasks.
%These tasks effectively evaluate the model's performance under different scenarios of classification, including balanced, imbalanced, and multi-class conditions.

 %This study used binary classification tasks for patient gender and abnormal blood potassium levels, as well as a 15-class classification task for arrhythmia detection. For  abnormal blood potassium levels, we matched the abnormal condition measured within one hour prior to the time of ECG recording. For arrhythmia detection, we selected the 14 most frequently occurring diagnoses, with the remaining ones grouped under "others," resulting in a total of 15 labels. Gender prediction is a binary classification task with a roughly balanced ratio of 50\% to 50\%, while the classification of normal vs. abnormal blood potassium levels has an imbalanced ratio of 97\% to 3\%. Additionally, there is a multi-class classification task for arrhythmia detection. These tasks effectively evaluate the model's performance under different scenarios of classification, including balanced, imbalanced, and multi-class conditions.

 % \subsection{Evaluated Models}
 \vspace{\baselineskip}

\noindent \textbf{Evaluated Models.} We select the following models for benchmarking: TimesNet~\cite{wu2022timesnet}, DLinear~\cite{zeng2023transformers}, GPT-2~\cite{radford2019language}, Llama 3.1~\cite{dubey2024llama}, MOMENT~\cite{goswami2024moment}, TEMPO~\cite{cao2023tempo} and ECG-FM~\cite{mckeen2024ecg}. The details of these models are shown in Table~\ref{pre_trained_models}. 
%\textit{ETT} is from electric power domain; \textit{Monash} contains 58 public time-series datasets with only one dataset from health domain which present patient counts related to medicala products; UCR/UEA contains 159 time-series datasets cover ECG. Both WebText and meta internal are pure textual dataset collected from Internet. 
For the TSDL, TSFM, and ECGFM model categories, the original data are downsampled to match the input length required by the pre-trained models. For the LLM, the input is designed as a prompt based on features calculated from the original ECG data.

 \begin{table}[h]
    \caption{Pre-Trained Datasets and Tasks of Benchmark Models. For datasets, only UCR/UEA, TSB-UAD, PhysioNet 2021, MIMIC-IV-ECG, UNH-ECG contains ECG data. For tasks, both waveform forecast and autoregressive use observed data to predict future time steps, masked time-series prediction can involve predict time-series masked out in the middle, while ECG-FM propose variants of masking tasks tailed for ECG.}
    \centering
    \resizebox{\textwidth}{!}{
        \begin{tabular}{c|c|c|c}
            \hline
            \textbf{Category} & \textbf{Models} & \textbf{Pre-Trained Dataset} & \textbf{Pre-Train Tasks} \\ \hline
            \multirow{2}{*}{TSDL} & TimesNet & ETT, Monash & waveform forecast \\ \cline{2-4}
                                  & DLinear & ETT, Monash & waveform forecast \\ \hline
            \multirow{2}{*}{LLM}  & GPT2 & WebText & autoregressive \\ \cline{2-4}
                                  & Llama3.1 & Meta internal corpus & autoregressive \\ \hline
            \multirow{2}{*}{TSFM} & MOMENT & ETT, Monash, UCR/UEA, TSB-UAD & masked time-series prediction \\ \cline{2-4}
                                  & TEMPO & ETT, Monash & waveform forecast \\ \hline
            ECGFM & ECG-FM & PhysioNet 2021, MIMIC-IV-ECG, UHN-ECG & wav2vec 2.0 Masking, CMSC, RLM \\ \hline
        \end{tabular}
    }
    \label{pre_trained_models}
\end{table}

\section{Results}

% The results are shown in Table~\ref{model_comparison}. 

% We denote RR Interval Estimation, Age Estimation, Gender Classification, Potassium Abnormality Prediction, and Arrhythmia Detection as RR., Age, Gen., Ka, AD, and denote zero-shot, few-shot, and fine-tune as zs, fs, and ft.

\begin{table}[h]
    \centering
    \small 
    \setlength{\tabcolsep}{3pt} 
    \renewcommand{\arraystretch}{0.9} 
    \caption{Benchmarking experimental results. Highlighted are the top \textcolor{teal}{first},  \textcolor{brown}{second},  and \textcolor{blue}{third} results. (RR Interval Estimation, Age Estimation, Gender Classification, Potassium Abnormality Prediction, Arrhythmia Detection, and zero-shot, few-shot, fine-tune are denoted as RR., Age, Gen., Ka, AD, and zs, fs, ft respectively.)}
    \resizebox{\textwidth}{!}{
        \begin{tabular}{m{1.1cm} c c|c|c|c|c|c|c|c@{}}
            \toprule
            & & & \textbf{TimesNet} & \textbf{DLinear} & \textbf{GPT2} & \textbf{LLama3.1} & \textbf{MOMENT} & \textbf{TEMPO} & \textbf{ECG-FM} \\ \midrule
            \multirow{6}{=}{\textbf{Regre. \newline (MAE)\(\downarrow\)}}  
            & \multirow{3}{*}{RR.} & zs & 817.0 ± 2.5 & 816.4 ± 2.9 & \textcolor{brown}{816.0 ± 2.5} & \textcolor{teal}{815.1 ± 1.9} & 816.6 ± 2.1 & \textcolor{blue}{816.3 ± 1.9} & 816.3 ± 1.9 \\ 
            &  & fs & 814.9 ± 1.9 & 816.0 ± 2.5 & 816.2 ± 1.2 & 816.2 ± 1.2 &  \textcolor{brown}{801.0 ± 1.4} & \textcolor{blue}{808.2 ± 2.6} & \textcolor{teal}{698.2 ± 96.7}  \\ 
            &  & ft & 304.3 ± 4.3 & 786.0 ± 5.4 & 823.1 ± 5.8 & 822.3 ± 3.1 & \textcolor{brown}{146.9 ± 1.3} & \textcolor{teal}{141.5 ± 2.1} & \textcolor{blue}{147.3 ± 1.3} \\ \cline{2-10}
            
            & \multirow{3}{*}{Age} & zs & \textcolor{brown}{62.28 ± 0.36} & 62.63 ± 0.50 & 62.40 ± 0.38 & 62.66 ± 0.14 & 62.61 ± 0.37 & \textcolor{blue}{62.33 ± 0.38} & \textcolor{teal}{62.27 ± 0.38} \\ 
            &  & fs & 62.58 ± 0.38 & 61.87 ± 0.32 & 62.30 ± 0.19 & 62.30 ± 0.19 & \textcolor{brown}{46.95 ± 2.16} & \textcolor{blue}{54.79 ± 1.64} & \textcolor{teal}{19.58 ± 5.95} \\ 
            &  & ft & 24.89 ± 0.07 & 28.46 ± 0.74 & 61.86 ± 0.56 & 61.61 ± 0.72 & \textcolor{teal}{13.41 ± 0.45} & \textcolor{blue}{13.52 ± 0.31} & \textcolor{brown}{13.49 ± 0.17} \\ \hline
    
            \multirow{6}{=}{\textbf{Binary Class \newline (F1 Score)\(\uparrow\)}}
            & \multirow{3}{*}{Gen.} & zs & \textcolor{teal}{0.60 ± 0.00} & \textcolor{brown}{0.51 ± 0.08} & 0.00 ± 0.00 & 0.20 ± 0.00 & 0.34 ± 0.00 & \textcolor{blue}{0.42 ± 0.00} & 0.33 ± 0.00 \\ 
            &  & fs & 0.33 ± 0.00 & \textcolor{brown}{0.49 ± 0.14} & 0.01 ± 0.00 & 0.04 ± 0.00 & \textcolor{teal}{0.52 ± 0.03} & \textcolor{blue}{0.36 ± 0.03} & 0.33 ± 0.00 \\ 
            &  & ft & 0.51 ± 0.05 & \textcolor{brown}{0.57 ± 0.01} & 0.04 ± 0.00 & 0.06 ± 0.00 & \textcolor{teal}{0.68 ± 0.02} & \textcolor{blue}{0.53 ± 0.01} & 0.34 ± 0.02 \\ \cline{2-10}
    
            & \multirow{3}{*}{Ka} & zs & \textcolor{blue}{0.06 ± 0.00} & 0.05 ± 0.00 & 0.00 ± 0.00 & 0.00 ± 0.00 & 0.02 ± 0.00 & \textcolor{brown}{0.18 ± 0.00} & \textcolor{teal}{0.35 ± 0.00}  \\ 
            &  & fs & 0.00 ± 0.00 & 0.06 ± 0.01 & 0.01 ± 0.00 & 0.01 ± 0.00 & \textcolor{teal}{0.49 ± 0.00} & \textcolor{teal}{0.49 ± 0.00} & \textcolor{brown}{0.34 ± 0.22} \\ 
            &  & ft & 0.01 ± 0.01 & 0.01 ± 0.01 & 0.01 ± 0.00 & 0.02 ± 0.00 & \textcolor{brown}{0.49 ± 0.00} & \textcolor{teal}{0.50 ± 0.00} & \textcolor{brown}{0.49 ± 0.00} \\ \hline
    
            \multirow{3}{=}{\textbf{15 Class \newline (ACC)\(\uparrow\)}} 
            & \multirow{3}{*}{AD} & zs & \textcolor{brown}{0.06 ± 0.01} & 0.03 ± 0.02 & 0.04 ± 0.00 & \textcolor{teal}{0.48 ± 0.01} & 0.03 ± 0.02 & 0.02 ± 0.00 & \textcolor{brown}{0.07 ± 0.00} \\ 
            &  & fs & 0.01 ± 0.00 & 0.40 ± 0.01 & \textcolor{brown}{0.48 ± 0.00} & \textcolor{brown}{0.48 ± 0.00} & \textcolor{teal}{0.49 ± 0.01} & 0.43 ± 0.07 & 0.18 ± 0.21 \\ 
            &  & ft & 0.03 ± 0.00 & 0.48 ± 0.02 & 0.49 ± 0.02 & 0.46 ± 0.06 & \textcolor{teal}{0.66 ± 0.03} & \textcolor{brown}{0.54 ± 0.14} & \textcolor{blue}{0.49 ± 0.03} \\ \hline

            \multirow{4}{=}{\textbf{Benchmark \newline (Win Rate)\(\uparrow\)}} & \multirow{4}{*}{Overall} & zs & 20\% & 0\% & 0\% & 40\% & 0\% & 0\% & 40\% \\
            & & fs & 0\% & 0\% & 0\% & 0\% & 60\% & 20\% & 20\%  \\
            & & ft & 0\% & 0\% & 0\% & 0\% & 60\% & 40\% & 0\%   \\
            & & ALL & 6.7\% & 0\% & 0\% & 13.3\% & \textcolor{teal}{40\%} & \textcolor{brown}{20\%} & \textcolor{brown}{20\%} \\
            \bottomrule
        \end{tabular}
    }
    \label{model_comparison}
\end{table}

\begin{figure}[h]
    \vspace{-0.3cm}
    \centering
    \includegraphics[width=\textwidth]{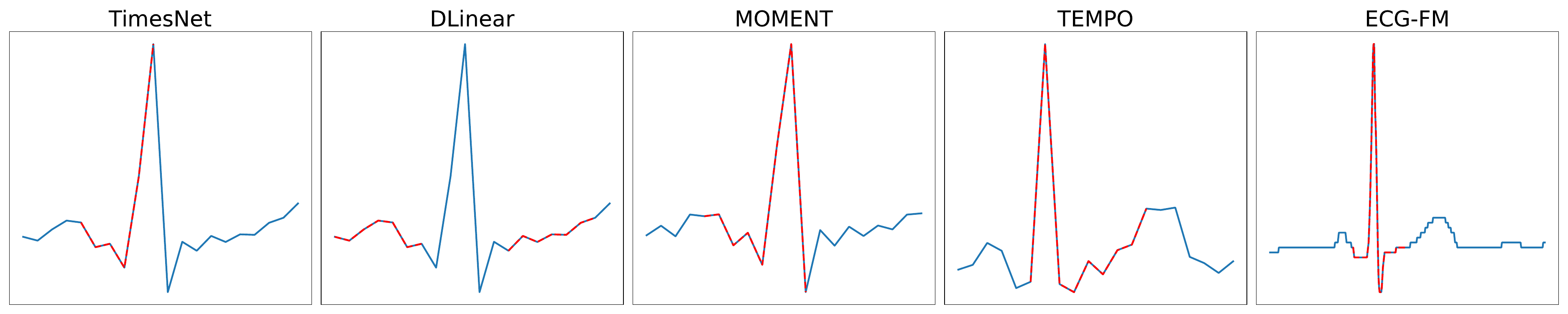}
    \caption{Saliency maps for the ECG-based RR Interval Estimation task. The blue line represents the ECG signal, the red line highlights the features the model focuses on. The same ECG segment is used, with downsampling applied due to varying input lengths of the pretrained models.}
    \vspace{-0.3cm}
    \label{map}
\end{figure}

As the performance and interpretation results presented in Table~\ref{model_comparison} and Figure~\ref{map}, respectively, the following observations can be made from the experimental results.

\ding{172} \textbf{Direct application of LLM on ECG is infeasible, showing inferior performance compared to TSDL.} LLMs struggled to determine gender and age from ECG, may due to insufficient knowledge linking demographic information to ECG features.

\ding{173} \textbf{Models pre-trained on time-series or ECG data outperform LLMs significantly.} LLMs are better suited for text tasks than time-series data processing. Extracting features for prompt design may lead to loss of crucial temporal information.

\ding{174} \textbf{Specialized ECGFM did not significantly outperform TSFM.} TSFM’s pretraining likely provides a robust understanding of time-series dynamics, enabling good adaptation to various tasks, compensating for the lack of ECG-specific training.

\ding{175} \textbf{Foundation model requires sufficient fine-tuning samples, as zero and few-shot performance was not good enough.} Differences between ECG data and pretraining data mean limited tuning may hinder effective understanding of ECG-specific tasks.

\ding{176} \textbf{Foundation model provides more interpretable results than TSDL.} Figure~\ref{map} shows the saliency maps for RR interval estimation. TSFM and ECG-FM effectively capture the feature peaks, demonstrating greater interpretability.

% (i) \textbf{\textit{The performance of the TSDL model on certain tasks degrades after tuning~(few-shot).}}
% (ii) \textbf{\textit{LLM perform poorly on downstream tasks involving ECG data.}}
% (iii) \textbf{\textit{Both TSFM and ECGFM perform well after tuning.}}
% (iv) \textbf{\textit{ECGFM does not demonstrate better performance in these five tasks.}}
% (v) MOMENT performs better in classification tasks.

%\input{sections/analysis}

\section{Discussion}

% To understand the relationship between the features extracted by each model when processing ECG data and their impact on the downstream tasks, we utilized saliency maps for visualization.
% As shown in Figure~\ref{map}, saliency maps provide interpretability by highlighting which parts of the ECG signal most influence the model's predictions, helping to understand the relationship between extracted features and downstream tasks.
% Here, we will examine the saliency maps of each model for the RR interval task, where the RR interval represents the time between two consecutive R-wave peaks in the ECG, indicating heart rate variability.
% From figure~\ref{map}, it is evident that, compared to TSDL (TimesNet and DLinear), TSFM and ECGFM place greater emphasis on the peaks, demonstrating a stronger capacity for feature understanding in these models. However, the excessive focus on peaks may lead to overfitting, thereby negatively impacting the performance of the models.

From the results presented in Table~\ref{model_comparison}, we can conclude that time-series foundation models are effective for ECG analysis. Based on the five summarized insights above, we discuss how these insights help us develop more advanced ECG foundation models in the future. From \ding{172}, applying LLM to ECG still needs specialized prompt design and external knowledge, which might require a retrieval-augmented generation (RAG) technique. Point \ding{173} addresses the importance of pre-training large-scale foundation models for various ECG downstream tasks. Moreover, based on \ding{174}, the pre-training data may include more than just ECG. The inclusion of general domain time-series could also boost the model performance on ECG. The current state-of-the-art TSFM and ECGFM still need amounts of fine-tuning samples as stated in \ding{175}. This motivates more efforts in the future to develop more advanced methods to pre-train or adapt the foundation model on ECG under zero and few-shots settings. From \ding{176}, the foundation model provides better interpretability compared to TSDL which paves a new way to explainable AI (XAI) in ECG analysis.

\section{Conclusion}

In this study, we build a comprehensive benchmark to evaluate various deep learning and foundation models for ECG analysis. Our results indicate that while time-series foundation models and ECG foundation models exhibit strong performance in certain tasks, suggesting their usefulness for ECG analysis, large language models struggle with ECG data, emphasizing the need for domain-specific and task-specific pre-training. Overall, our findings highlight the strengths and limitations of different foundation models for ECG analysis, underscoring the importance of foundation models and robust benchmarks for them.

\section{Acknowledgement}

This research was partially supported by the US National Science Foundation under Award Number 2319449 and Award Number 2312502, as well as the US National Institute of Diabetes and Digestive and Kidney Diseases of the US National Institutes of Health under Award Number K25DK135913.

\bibliographystyle{unsrt}
\bibliography{ref}

\end{document}